\documentclass[conference]{IEEEtran}
\IEEEoverridecommandlockouts
\usepackage{cite}
\usepackage{amsmath,amssymb,amsfonts}
\usepackage{algorithmic}
\usepackage{graphicx}
\usepackage{textcomp}
\usepackage{xcolor}
\usepackage{multirow}
\usepackage[font=footnotesize]{caption}
\usepackage{subcaption}
\usepackage{comment}
\DeclareMathOperator{\E}{\mathbb{E}}
\def\BibTeX{{\rm B\kern-.05em{\sc i\kern-.025em b}\kern-.08em
    T\kern-.1667em\lower.7ex\hbox{E}\kern-.125emX}}
\begin{document}

\title{Identity-Preserving Realistic Talking Face Generation}

\author{\IEEEauthorblockN{
Sanjana Sinha, Sandika Biswas and
Brojeshwar Bhowmick}
\IEEEauthorblockA{Embedded Systems and Robotics,\\
TCS Research and Innovation\\
Email: \{{sanjana.sinha, biswas.sandika, b.bhowmick\}}@tcs.com
}
}

\maketitle

\begin{abstract}

Speech-driven facial animation is useful for a variety of applications such as telepresence, chatbots, etc. The necessary attributes of having a realistic face animation are 1) audio-visual synchronization (2) identity preservation of the target individual (3) plausible mouth movements (4) presence of natural eye blinks. The existing methods mostly address the audio-visual lip synchronization, and few recent works have addressed synthesis of natural eye blinks for overall video realism. In this paper, we propose a method for identity-preserving realistic facial animation from speech. We first generate person-independent facial landmarks from audio using DeepSpeech features for invariance to different voices, accents, etc. To add realism, we impose eye blinks on facial landmarks using unsupervised learning and retarget the person-independent landmarks to person-specific landmarks to preserve the identity-related facial structure which helps in generation of plausible mouth shapes of the target identity.  Finally, we use LSGAN to generate the facial texture from person-specific facial landmarks, using an attention mechanism that helps to preserve identity-related texture. An extensive comparison of our proposed method with the current state-of-the-art methods demonstrate a significant improvement in terms of lip synchronization accuracy,  image reconstruction quality,  sharpness, and identity-preservation. A user study also reveals improved realism of our animation results over the state-of-the-art methods. To the best of our knowledge, this is the first work in speech-driven 2D facial animation that simultaneously addresses all the above-mentioned attributes of a realistic speech driven face animation.
\end{abstract}

\begin{IEEEkeywords}
Talking face, motion-texture decoupling, realistic face animation, identity preservation.  
\end{IEEEkeywords}

\section{Introduction}
\label{section:Introduction}
Generating a realistic talking face from speech input is a fundamental problem with several applications such as virtual reality, computer-generated imagery (CGI), chatbots, telepresence, etc.  Essential requirements for all the applications are that the synthesized face must appear photo-realistic with accurate and realistic audio-visual lip synchronization, and must also preserve the identity of the target individual. 
Also, for most of these applications, it is expected to have a single image with the target identity's face on which the motion has to be induced from a given speech as input, for greater flexibility of changing the target subjects at test time. Hence, audio-driven realistic facial animation from a single image input is crucial.
In general, any speech-driven facial animation method has several challenges due to the existence of a variety in the facial structures of different target identities, different voices, and accents in input audio, etc.

\begin{figure}[t!]
    
    \begin{subfigure}[t]{0.5\columnwidth}
        \centering
        \includegraphics[trim=1cm 1cm 0.9cm 1cm, width=\linewidth]{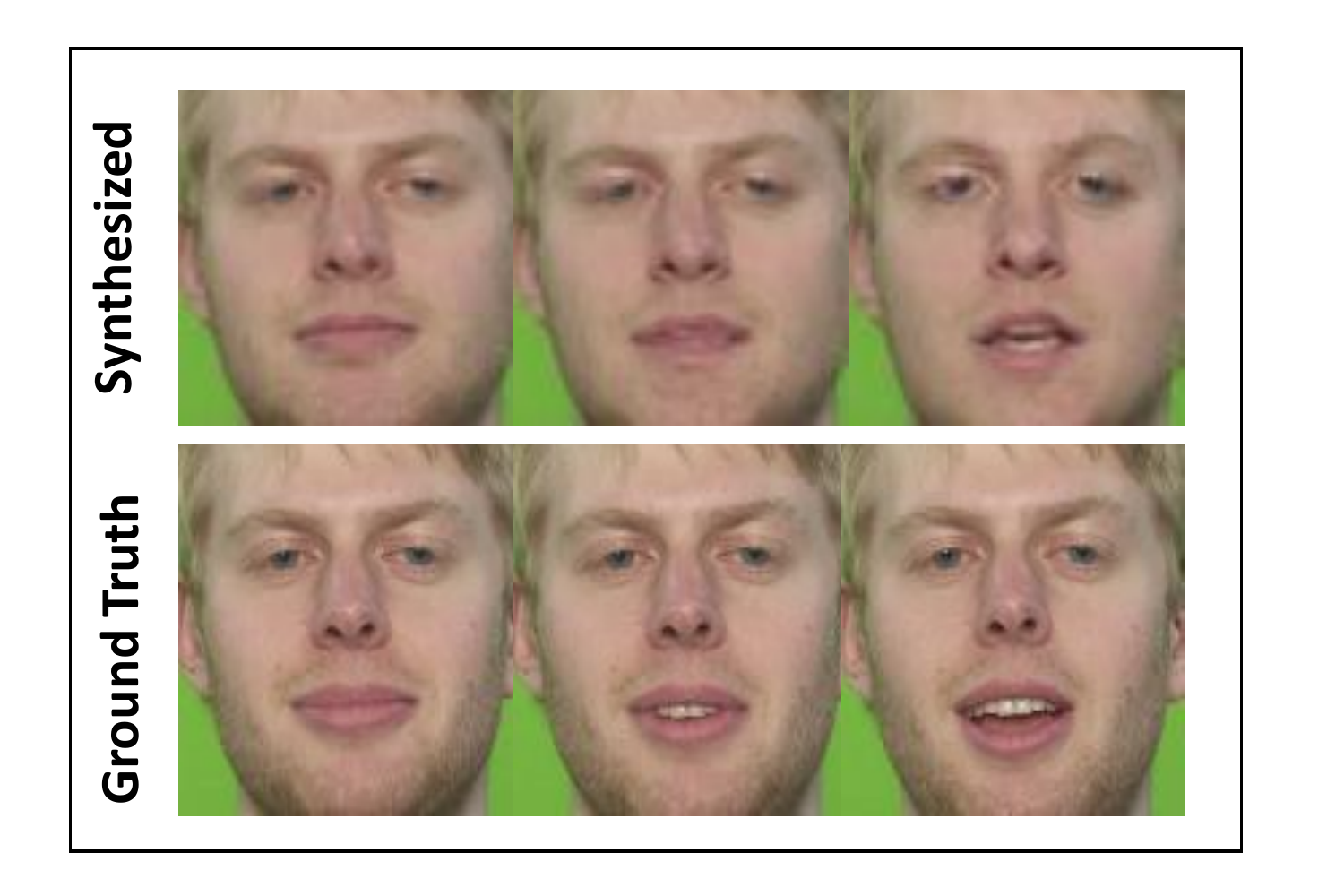}
        \caption{}
        \label{subfig:identity}
    \end{subfigure}%
    ~
    \begin{subfigure}[t]{0.5\columnwidth}
        \centering
        \includegraphics[trim=0.7cm 1cm 1cm 1cm, width=\linewidth]{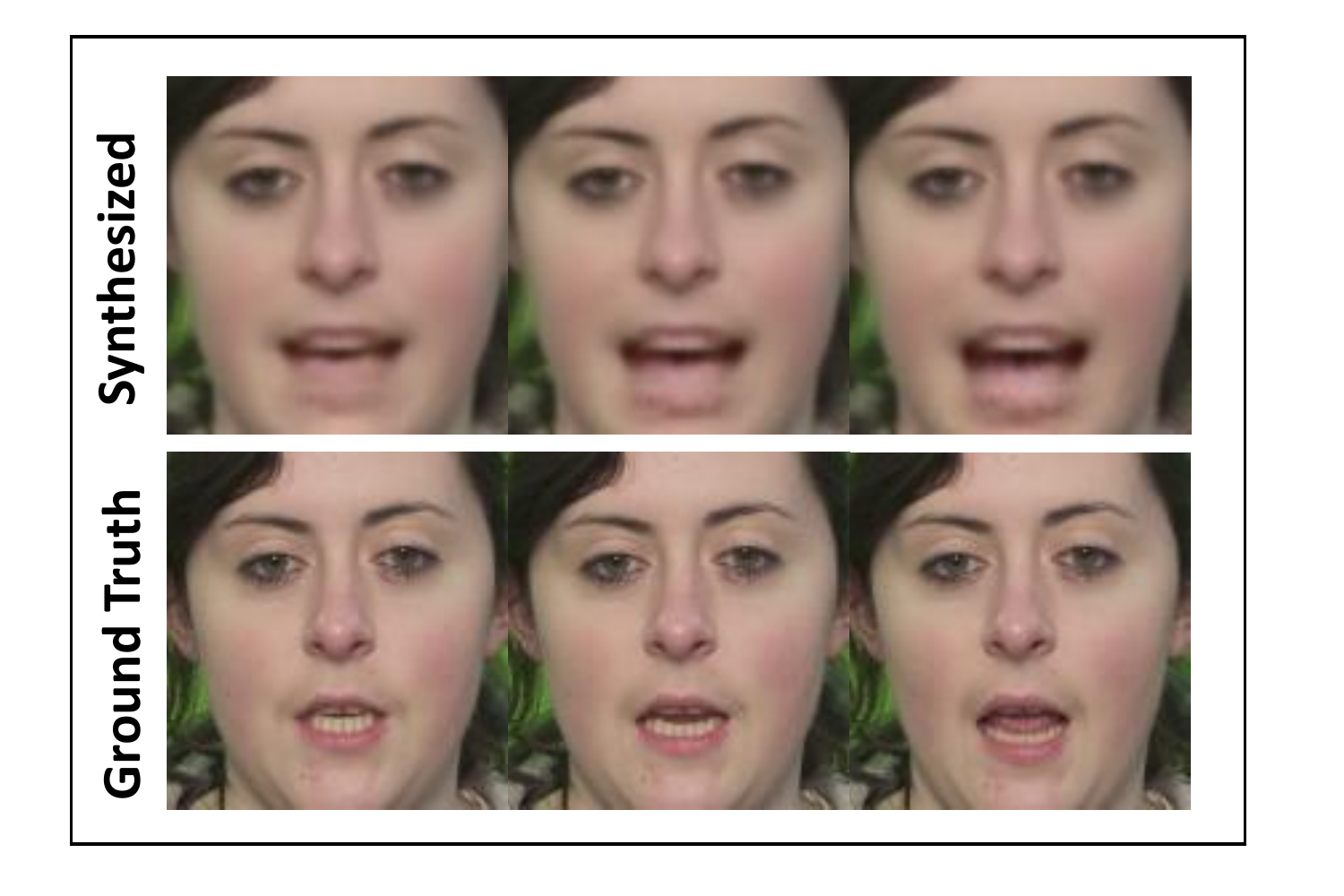}
        \caption{}
        \label{subfig:shape}
    \end{subfigure}
    \caption{ Issues with current methods in 2D facial animation (a) Difference in image texture of synthesized face produced by Vougioukas et al. \cite{vougioukas2019realistic} from the ground truth image texture leads to perceived difference in identity of the rendered face from the target individual. (b) Despite synchronization with audio, the facial animation sequence synthesized using the method of Chen et al. \cite{chen2019hierarchical} contains implausible or unnatural mouth shape (last frame) that can be perceived as being fake. The results were obtained by evaluation using pre-trained models made publicly available by the respective authors. }
\label{fig:identity_realistic}
\end{figure}
\begin{figure*}[t]
\center
\includegraphics[ width=\textwidth]{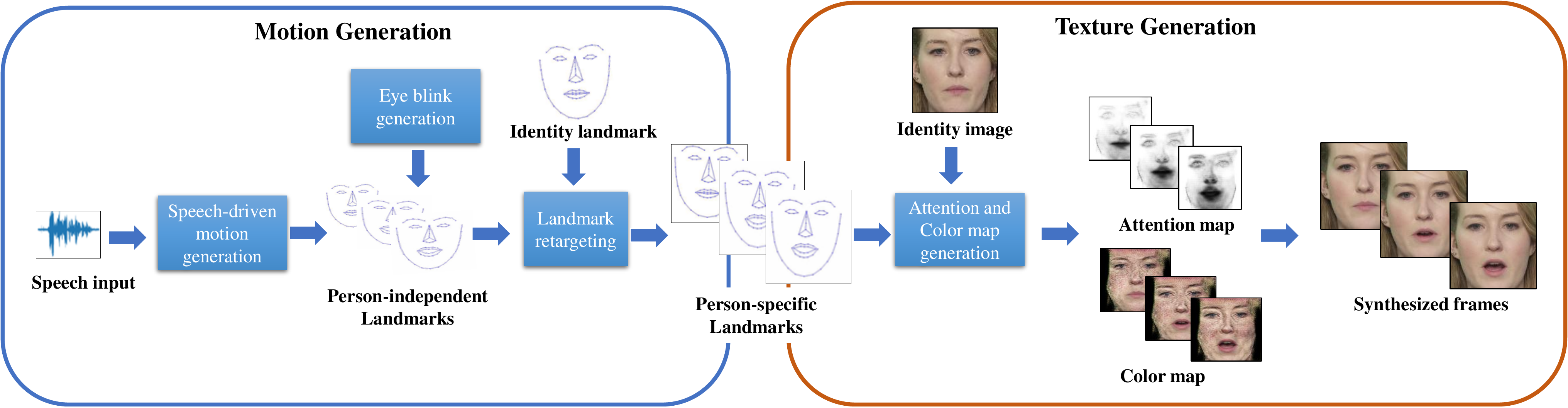}

\caption{Our proposed method consists of the following 4 stages - (1) speech driven motion generation on person-independent landmarks, (2) eye-blink generation (3) retargeting of generated motion on person-specific landmarks, (4) Synthesizing face images using attention and color map generation. Attention generation helps segregate identity information, represented by lighter regions of attention map, from motion-based texture (darker regions).} 
\label{fig:concept_diagram}
\end{figure*}

Most of the existing methods for facial video synthesis \cite{chung2017you,chen2019hierarchical,zhu2018high,zhou2019talking,song2018talking} focus on generating facial movements synchronized with speech, while only a few \cite{vougioukas2019end,vougioukas2019realistic} have addressed the generation of spontaneous facial gestures such as eye blinks that add realism to the synthesized video. However, the latest methods either fail to preserve the perceived identity of the target individual (Fig. \ref{subfig:identity}), or generate implausible or unnatural shape of the mouth in a talking face (Fig. \ref{subfig:shape}). 
Lack of resemblance with given identity or change of identity in consecutive synthesized frames (Fig. \ref{subfig:identity}) can give rise to the  
uncanny valley effect \cite{mori2012uncanny}, in which the facial animation can be perceived as visually displeasing or eerie to the viewer. 
Moreover, the lack of any natural and spontaneous movements over the talking face except around the mouth region can be an indication of synthesized videos.

In this paper, we address the above issues for generating realistic facial animation from speech. To the best of our knowledge, this is the first work on speech-driven 2D facial animation which simultaneously addresses the following attributes required for realistic face animation: 1) audio-visual synchronization (2) identity-preserving facial texture (3) generation of plausible mouth movements  (4) presence of natural eye blink movements.   
Inspired by a recent method \cite{chen2019hierarchical}, we first generate a high-level representation of the face using 2D facial landmarks to capture the motion from speech, then use an adversarial network for generating texture by learning motion-based image attention. Our approach is outlined in Fig. \ref{fig:concept_diagram}.
\begin{figure}[t]
\center
\includegraphics[width=\columnwidth]{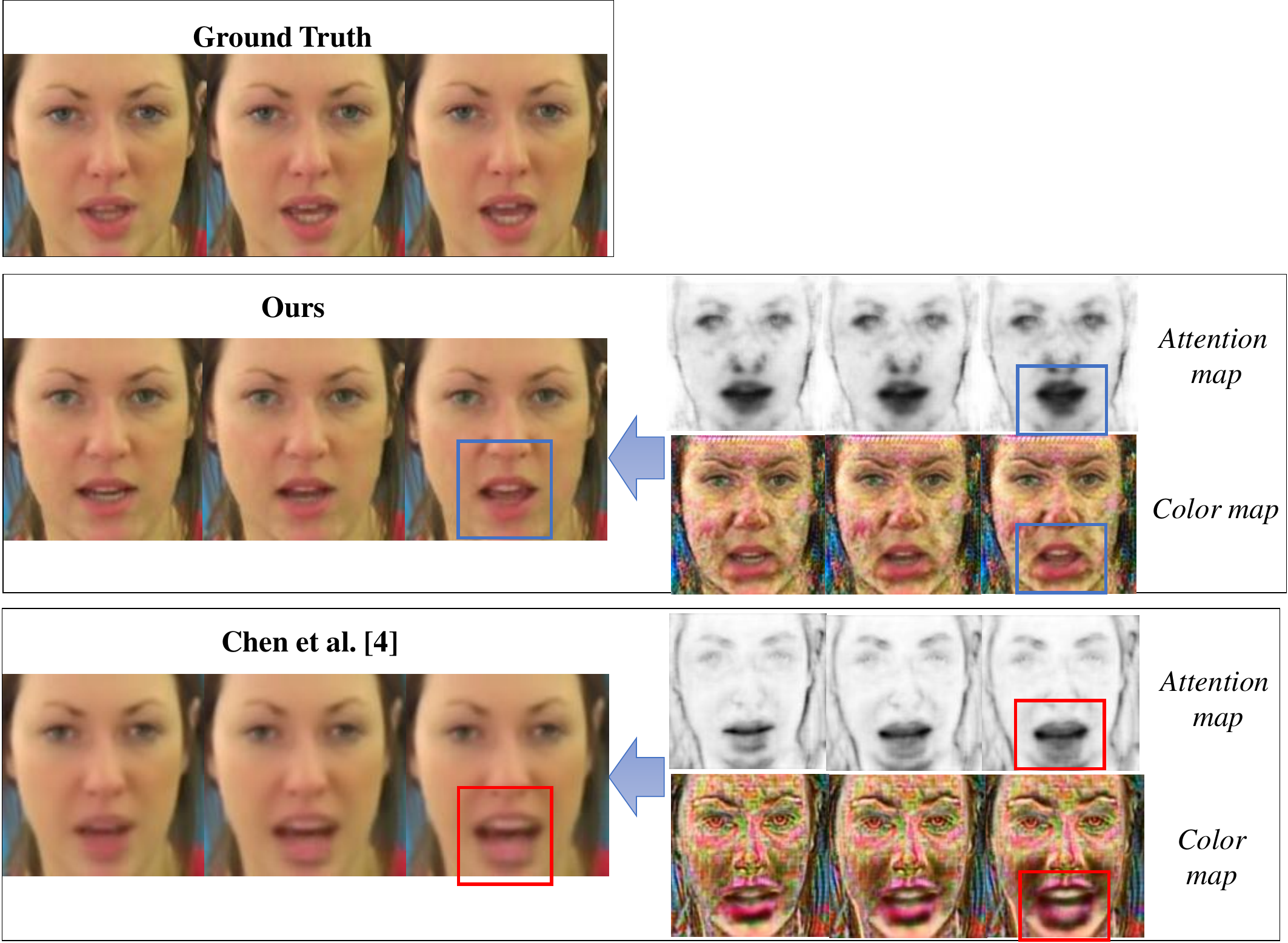}

\caption{Effect of intermediate attention and color map on the final texture.  Intermediate attention values (grey areas) of extended regions surrounding the lips in the attention map $^\dagger$ generated by Chen et al. \cite{chen2019hierarchical} (last row) results in the blurred texture and unusual shape of the mouth in the animated face (last frame). Whereas uniformly low attention values (dark areas) in the mouth region in our attention map and distinct lip shape and texture in our color map leads to generation of sharp facial texture with plausible shape of the mouth.
$^\dagger$ Actual attention map (where higher values indicate regions with more significant motion) generated by \cite{chen2019hierarchical} is inverted here for direct comparison with our attention map (lower values indicate regions with more significant motion).} 
\label{fig:comparewithSOA}
\end{figure}
The challenge is the decoupling of speech-driven motion from identity-related attributes such as different facial structures, face shapes, etc. for robust motion prediction. To address this, we first learn speech-related motion on identity-independent landmarks. Then, the learnt landmark motion is transferred to the person-specific landmarks for generating identity specific facial movements.
Unlike state-of-the-art methods for speech-driven 2D facial animation, we use DeepSpeech \cite{hannun2014deep} features of given audio input, which exhibits greater robustness to the variety in audio that exists due to different audio sources, accents, and noise.
Since eye blinks are unrelated to speech, we generate blink motion independently from audio-related landmark motion.
Finally, we learn an attention map and color map from the identity image and the predicted person-specific landmarks. The attention map \cite{pumarola2018ganimation} helps in segregating regions of facial motion (defined by the lower values of attention) from the rest of the face containing identity-related information (defined by higher values of attention). The color map contains a novel texture for the facial regions where the attention map indicates motion. We use the combination of attention map and color map to generate the final texture. Texture in regions of motion is obtained from the color map, while the texture in the rest of the face is obtained from the input identity image (driven by the weights of the attention map). Our network learns the attention map and the color map without explicit attention or color map labels for supervision. 

The quality of the learned attention map is extremely crucial for the overall quality of the generated face. Fig. \ref{fig:comparewithSOA} shows an example of synthesized face images by Chen et al. \cite{chen2019hierarchical} where the final texture of the animated face is adversely affected by the values of intermediate attention map and color map. In regions of facial motion surrounding the mouth, uniform regions of very low values (dark regions) of the attention map are needed for sharp texture generation, while intermediate values (grey regions) lead to blur in mouth texture (shown in Fig \ref{fig:comparewithSOA} last row). In regions of low attention (dark regions of the attention map indicating motion), the color map values contribute to the overall sharpness of the generated texture and the shape of the mouth.
To address the problem of accurate attention and color map generation, we propose an architecture for texture generation which uses LSGAN \cite{mao2017least} for learning sharp image texture and plausible mouth shapes (Fig. \ref{fig:comparewithSOA} second row). Moreover, during adversarial training, if attention values become very low in static facial regions,
it can lead to texture blur and also possible loss of identity information. 
Hence, regularization is also needed as an additional constraint in the learning of the attention map.  
Unlike Chen et al. \cite{chen2019hierarchical}, we use spatial and temporal $L_2$ regularization on the attention and color map for generating smooth motion and plausible mouth shapes without loss of identity.

\noindent
The main contributions of our paper are:
\begin{itemize}
    \item We propose a four-stage approach for speech-driven 2D face synthesis that helps to achieve realistic facial animation which contains plausible mouth movements synchronized with speech, natural eye blinks, and preserves the identity information of the target subject.
    \item Our proposed method generates an intermediate landmark representation that defines the speech-induced facial motion along with realistic eye blinks. Further, this intermediate representation is used to generate the facial texture with motion defined at the landmark stage.
    \item Our proposed audio-to-landmark generator network uses DeepSpeech features to learn motion on facial landmarks for better generalization to new voices, accents, and noise. 
    \item We carry out unsupervised learning of eye blinks on facial landmarks using MMD loss minimization. 
    \item Our texture generation network produces identity-preserving facial texture from identity-specific facial landmarks using a combination of attention generation, attention regularization, and least-squares adversarial training. 
\end{itemize}

\section{Related Work}

\textbf{Talking face generation:} 
Generating realistic talking faces from audio has been a research problem in the computer vision and graphics community for decades \cite{yehia2002linking,cao2005expressive,simons1990generation}. 
Recent research works have carried out the speech-driven 
synthesis of lip movements \cite{chen2018lip}, as well as animation of the entire face in 2D \cite{chung2017you,chen2019hierarchical,vougioukas2019end,vougioukas2019realistic,zhu2018high,zhou2019talking,song2018talking}.  Earlier approaches have carried out subject-specific talking face synthesis from speech \cite{suwajanakorn2017synthesizing,garrido2015vdub,fan2015photo}. However, these approaches require a large amount of training data of the target subject, and such subject-specific models cannot generalize to a new person. Subject-independent facial animation was carried out by \cite{chung2017you} from speech audio and a few still images of the target face. However, the generated images contain blur due to $L_{1}$ loss minimization on pixel values and an additional de-blurring step is required. On the other hand, Generative Adversarial Networks (GANs) \cite{goodfellow2014generative} are widely used for image generation due to their ability to generate sharper, more detailed images compared to networks trained with only $L_{1}$ loss minimization. Recent GAN-based methods \cite{chen2017deep,chen2019hierarchical,vougioukas2019realistic,vougioukas2019end,zhou2019talking,song2018talking} have generated facial animation from arbitrary input audio and a single image of the target identity. In this work, we adopt a GAN-based approach for synthesizing face images from the motion of intermediate facial landmarks, which are generated from audio.

\textbf{Talking face with realistic expression:}
Current methods \cite{chen2017deep,chen2019hierarchical,zhou2019talking,song2018talking} have mostly addressed audio-synchronization instead of focusing on overall realism of the rendered face video. The absence of spontaneous movements, such as eye blinks can also be an indication of synthesized videos \cite{li2018ictu}.  Few works \cite{vougioukas2019end,vougioukas2019realistic} have addressed this problem by using adversarial learning of spontaneous facial gestures such as blinks. 
However, these methods generate facial texture without the use of landmark-guided image attention, which can lead to loss of facial identity (Fig. \ref{subfig:identity}). In this work, inspired by \cite{vougioukas2019realistic} we perform eye blink generation for the realism of synthesized face videos. Unlike \cite{vougioukas2019realistic}, the blink motion is generated on facial landmarks to ensure decoupled learning of motion and texture for better identity preservation. 

\textbf{Segregation of motion from texture:}
In talking face synthesis, subject-related and speech-related information are separately addressed in \cite{zhou2019talking} by learning disentangled audio-visual information, i.e., complementary representations for speech and identity, thereby generating talking face from either video or  speech. Using high-level image representations such as facial landmarks \cite{kazemi2014one} is another way to segregate speech-related motion from texture elements such as identity information, viewing angle, head pose, background, illumination. A recent research work \cite{chen2019hierarchical} adopts a two-stage approach in which facial motion is decoupled from texture using facial landmarks. 
Although we also use facial landmarks to segregate motion from texture, unlike \cite{chen2019hierarchical}, our approach involves 
imposing natural facial movements like eye blinks in addition to lip synchronization with given audio input. We retarget the person-independent landmarks with audio-related motion and blinks to person-specific landmarks for subsequent texture generation. This helps in generating plausible mouth shapes in the target facial structures.


\section{Our Approach}

\begin{figure}[t]
\center
\includegraphics[width=0.8\columnwidth]{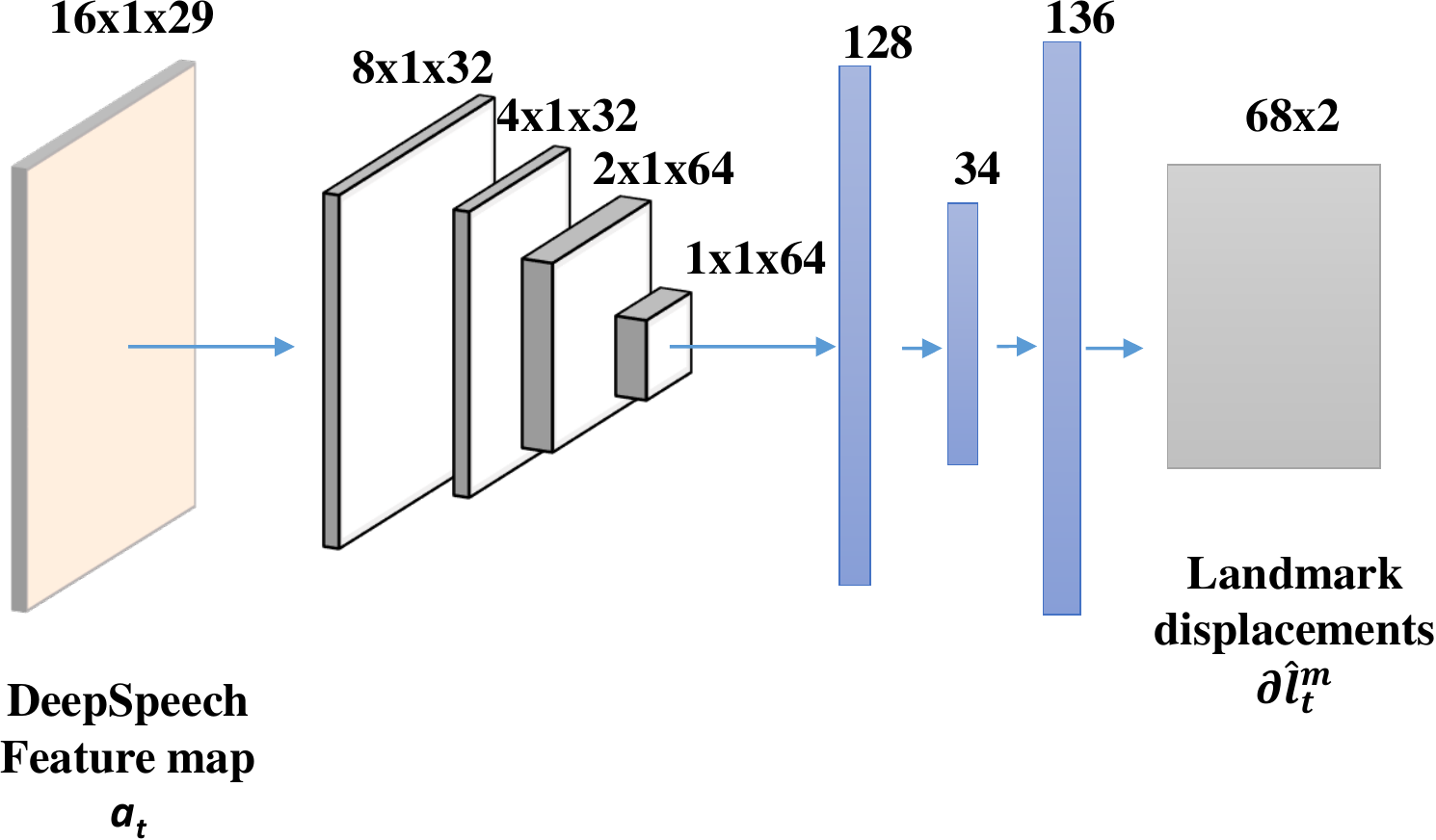}
\caption{Network architecture for audio-to-landmark prediction.}
\label{fig:aud-to-land}
\end{figure}

\subsection{Speech-driven Motion Prediction}
For a given speech signal represented by a sequence of overlapping audio windows $ A=\{A_0,A_1 \cdots A_t\} $, we first predict the speech-induced motion on a sparse representation of the face $ l^p=\{l^p_0,l^p_1 \cdots l^p_t \}$ where $l^p_t \in \mathbb{R}^{68 \times 2}$ consists of 68 facial landmark points representing eyes, eyebrows, nose, lips, and jaw. Unlike the state-of-the-art methods, we use DeepSpeech features \cite{hannun2014deep} instead of using audio MFCC features. DeepSpeech features are used for gaining robustness against noise and invariance to audio input from a variety of speakers. DeepSpeech features $ a=\{a_0, a_1, \cdots a_t \}$ where $a_t \in \mathbb{R}^{16 \times 29}$, corresponding to audio windows $ A=\{A_0,A_1 \cdots A_t\}$ are used for landmark generation. 


\textbf{Landmark prediction from speech:}
Facial landmarks for different subjects contain person-specific facial attributes i.e., different face structures, sizes, shapes, and different head positions. Speech driven lip movements for a given audio segment are independent of these variations. So to make landmark prediction invariant to these factors, we consider a canonical landmark representation $ l^m=\{l^m_0,l^m_1 \cdots l^m_t \};$ where, $l^m_t \in \mathbb{R}^{68 \times 2}$, which is mean of facial landmarks over the entire dataset. We consider a frontal face with closed lips as the neutral mean face, $l^m_N$. 
We train our speech-to-landmark generation network to predict $ \delta l^m=\{\delta l^m_0,\delta l^m_1 \cdots \delta l^m_t \}$ where, $\delta l^m_t \in \mathbb{R}^{68 \times 2}$ represents displacement from the neutral mean face $l^m_N$. 
 Person-specific facial landmarks $l^p_t$ is calculated from canonical landmark displacements $ \delta l^m_t$ from $l^m_N$ using,
\begin{equation}
l^p_t = \delta l^m_t * S_t + PA(l^p_N,l^m_N)
\end{equation}
where, $PA(l^p_N,l^m_N)$ represents the rigid Procrustes alignment \cite{srivastava2005statistical} of $l^m_N$ with $l^p_N$ as reference. $S_t$ represents scaling factor (ratio of height and width of person-specific face to mean face). $\delta l^m_t * S_t$ represents displacements of person-specific landmarks $\delta l^p_t$.

The network is trained with full supervision ($L_{lmark}$) for a one-to-one mapping of DeepSpeech features to landmark displacements. 
\begin{equation}
L_{lmark}=||\delta l^m_t-\hat{\delta l^m_t}||^2_2
\end{equation}
$\delta l^m_t$ and $\delta \hat{l^m_t}$ represents ground-truth and predicted canonical landmarks displacements.

A temporal loss ($L_{temp}$) is also used to ensure consistent displacements over consecutive frames as present in ground truth landmark displacements. 
\begin{equation}
L_{temp}=||(\delta l^m_t-\delta l^m_{t-1})-(\hat{\delta l^m_t}-\hat{\delta l^m}_{t-1})||^2_2
\end{equation}
Total loss ($L_{tot}$) for landmark prediction is defined as,
\begin{equation}
L_{tot}=\lambda_{lmark}L_{lmark}+\lambda_{temp}L_{temp}
\end{equation}
where $\lambda_{lmark}$ and $\lambda_{temp}$ defines weightage of each of the losses. 
\begin{figure}[t]
\center
\includegraphics[width=0.8\columnwidth]{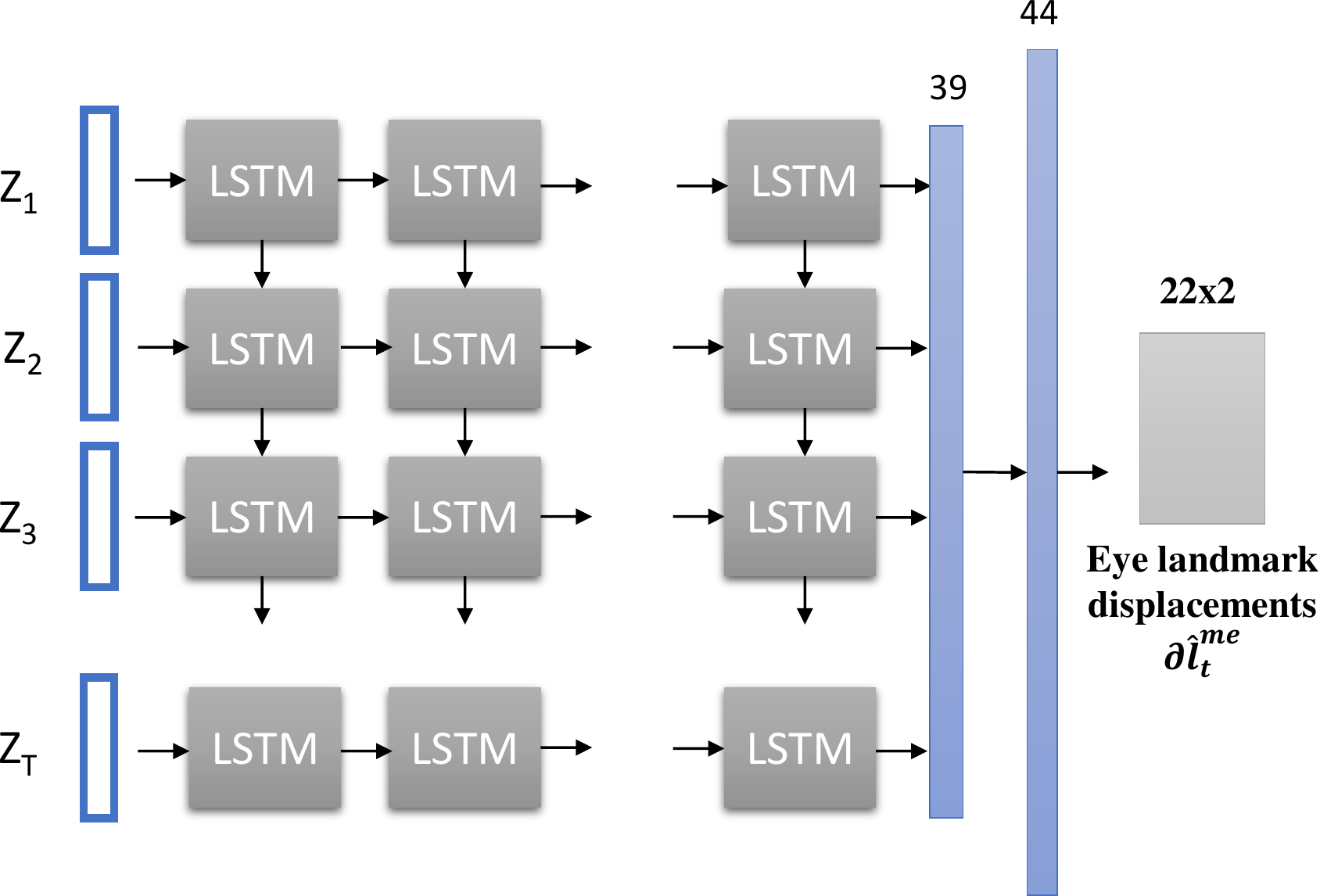}
\caption{Network architecture for blink generation.}
\label{fig:blink-generation}
\end{figure}

\subsection{Spontaneous Blink Generation}

Unlike previous approaches which use landmarks for facial animation \cite{chen2019hierarchical}, we impose eye blinks on the facial landmarks for adding realism to facial animation. Unlike end-to-end methods that generate natural facial expressions and eye blinks \cite{vougioukas2019end}, our blink movements are learnt over the sparse landmark representation for better preservation of identity-related texture. 

We train the blink generation network to learn a realistic eye blink, duration of eye blinks, and permissible intervals between two blinks from the training datasets. As there is no dependency of blinks on speech input, we generate eye blinks in an unsupervised manner only from random noise input. 
We aim to learn blink patterns, blink frequencies, and blink duration over the training dataset via unsupervised learning. In literature, generative adversarial networks (GAN) \cite{goodfellow2014generative} have been used for image generation from random noise input. Training of GAN requires optimization of a min-max problem, which is often difficult to stabilize. Li et al. \cite{li2015generative} have proposed a simpler category of GAN where the discriminator is replaced with a straightforward loss function that matches different moments of ground-truth (real) and predicted (fake) distributions using maximum mean discrepancy (MMD) \cite{gretton2007kernel}. 

We use MMD loss $L_{MMD^2}$ to match distribution of each landmark displacements over a sequence length $T$ .
\begin{multline}
L_{MMD^2}=\frac{1}{N^2}\sum_{i=1}^{N}\sum_{i'=1}^{N}k(\delta l_i^{me},\delta l_{i'}^{me}) \\ -\frac{2}{NM}\sum_{i=1}^N\sum_{j=1}^Mk(\delta l_i^{me},\hat{\delta l}_j^{me}) -\frac{1}{M^2}\sum_{j=1}^{M}\sum_{j'=1}^{M}k(\hat{\delta l}_j^{me},\hat{\delta l}_{j'}^{me})
\end{multline}


\noindent
where, $k(x,y)=exp(-\frac{|x-y|^2}{2\sigma})$ is used as the kernel for comparing the real and fake distributions. $\delta l^{me}$ and ${\hat{\delta l}}^{me}$ represents ground truth and predicted distribution of displacements of each of the landmark points in eye region over sequence $T$. We also use min-max regularization on predicted distributions to enforce it to be within the range of average displacements seen in the training dataset.

\subsection{Attention-based Texture Generation}
\begin{figure*}[t]
\centering
\includegraphics[width=0.7\linewidth]{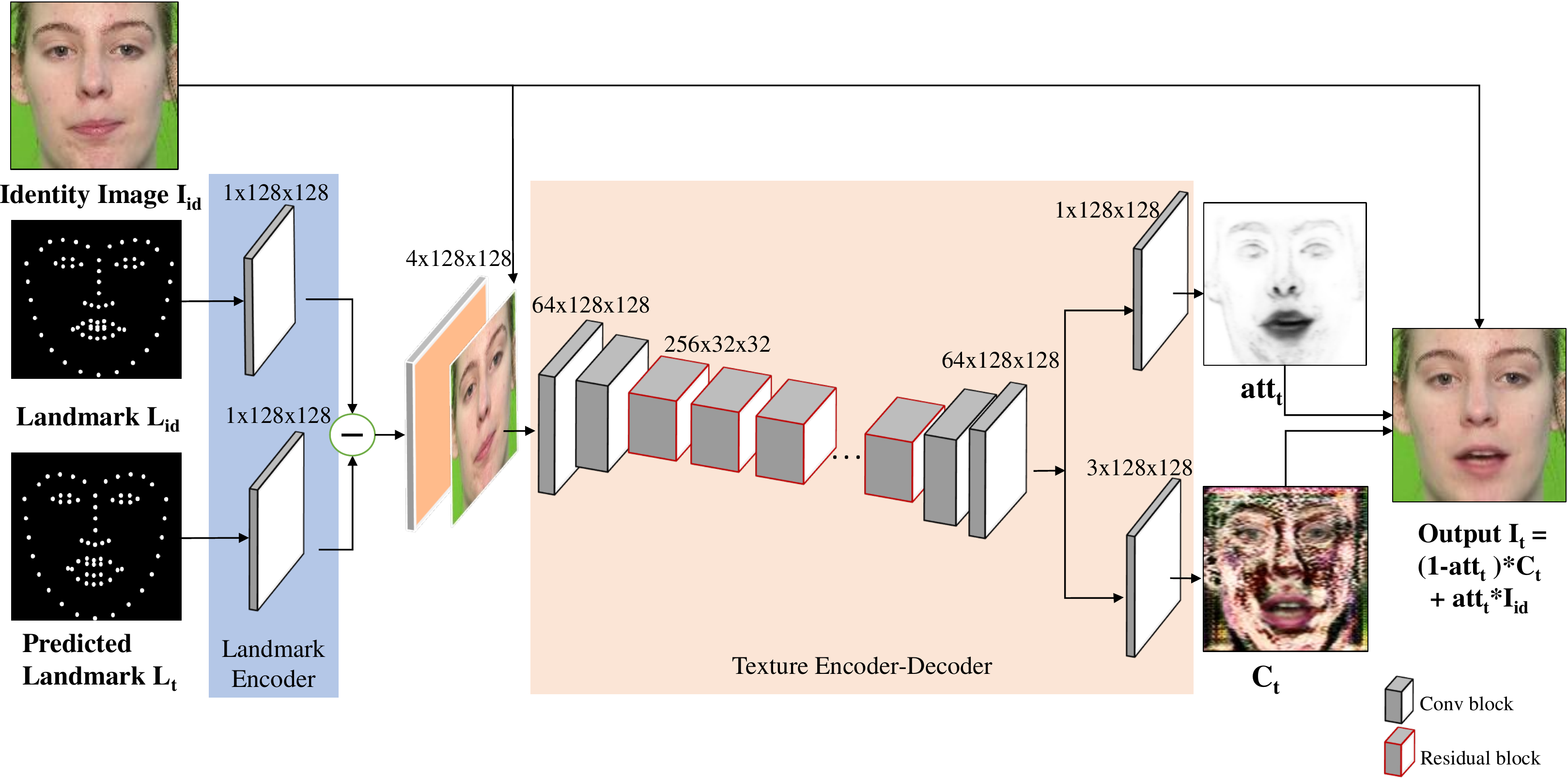}
\caption{The architecture of our proposed texture generator network.}
\label{fig:architecture}
\end{figure*}

Given a single image of the target identity $I_{id}$, the objective is to transform a sequence of person-specific facial landmarks $l^p=\{l^p_0,l^p_1 \cdots l^p_t\}$ into a sequence of photo-realistic images $I=\{\mathit{I_0},\mathit{I_1} \cdots \mathit{I_t}\}$ that accurately reflect the facial expressions corresponding to the input landmark images $L$ (image representation of the $68 \times 2$ landmarks $l^p$). A generative adversarial network is trained using ground truth video frames $I^*$ and the corresponding ground-truth landmark images $L^*$ . Since the texture generation network is trained on ground-truth landmarks, the network learns to generate face texture for eye blinks. During evaluation, the predicted speech-driven landmarks with imposed eye blinks are used as input for texture generation.

Our generator network focuses on generating novel texture for image regions that are responsible for facial expressions (defined by motion on landmarks), while retaining texture from $I_{id}$ in the rest of the image. This is achieved by learning a grayscale attention map and an RGB color map over the face image instead of directly regressing the entire face image, using a similar approach presented in \cite{pumarola2018ganimation,chen2019hierarchical}. The attention map $att_t$ determines how much of the original texture values in $I_{id}$ will be present in the final generated image ${I_t}$. The color map $C_t$ contains the novel texture in the regions of facial motion. 

The final generated image $I_{t}$  is derived as follows:
\begin{equation}
    I_{t} = (1-att_t) * C_t + att_t * I_{id}
\end{equation}




The network is trained by minimizing the following loss functions:

\textit{Pixel Intensity loss} : This is a supervised loss on the RGB intensity values of the entire image with a special emphasis on the eyes and mouth regions.
\begin{equation}
    L_{pix} = \sum_{t} \alpha |I_{t} - I^*_{t}|
\label{pixel_loss}
\end{equation}
where, $\alpha$ represents a fixed spatial mask representing weights assigned to individual pixels for contributing to the overall loss, with higher weights assigned to the regions surrounding the mouth and eyes. A fixed $\alpha$ has been experimentally found to be more stable than a dynamic pixel mask dependent on $att_t$ used in \cite{chen2019hierarchical}. 

\textit{Adversarial loss}: Using only the pixel intensity loss $L_{pix}$ results in a considerable blur in the generated image due to the $L_{1}$ distance minimization. A discriminator network is used to make the generated texture sharper and more distinct, especially in regions of motion. We adopt the LSGAN \cite{mao2017least} for adversarial training of our texture generation network, because of its better training stability as well as its ability to generate higher quality images than the regular GAN. Regular GANs use the sigmoid cross-entropy loss function, which is prone to the problem of vanishing gradients, in which the gradient becomes small for generated images that lie far from the decision boundary. 
The LSGAN helps overcome this problem by using the least-squares loss function which penalizes samples that are correctly classified yet far from the decision boundary. Due to this property of LSGANs, the generation of samples is closer to real data \cite{mao2017least}. The LSGAN loss functions for the discriminator and generator are :
\begin{align}
     L(D) &= \frac{1}{2} \E_{x \sim p_{I}(x)}[(D(x)-1)^2] + \frac{1}{2} \E_{z \sim p_{z}(z)}[D(G(z))^2]
    &&\\
    L(G)  &= \frac{1}{2} \E_{z \sim p_{z}(z)}[(D(G(z))-1)^2]&&
\end{align}
where $p_{I}$ is the distribution of the real face images and $p_{z}$ is the distribution of the latent variable $z$.
The adversarial loss $L_{adv}$ is computed as follows:
\begin{equation}
L_{adv} = L(G) + L(D)
\end{equation}

\textit{Regularization loss} : No ground-truth annotation is available for training the attention map and color map. Low values of the attention map in the regions of the face other than the regions of motion would result in blurring of the generated texture. Hence, a $L_{2}$ regularization is applied to prevent the attention map values from becoming too low. 
\begin{equation}
    L_{att} = \sum_{t} ||1-att_t||_2
\end{equation}
To ensure the continuity in the generated images, a temporal regularization is also applied by minimizing first-order temporal differences of attention and color maps.
\begin{equation}
L_{temp} = \sum_{t} ||(att_t - att_{t-1})||_2 + \sum_{t} ||(C_t - C_{t-1})||_2
\end{equation}
The total regularization loss is :
\begin{equation}
    L_{reg} = L_{att} + L_{temp}
\end{equation}
The final objective function of generator is to minimize the following combined loss:
\begin{equation}
L = \lambda_{pix} L_{pix} + \lambda_{adv}L_{adv} 
       + \lambda_{reg} L_{reg} 
       \label{eqn:textureobjectivefunction}
\end{equation}
\label{sec:III-C}
where, $\lambda_{pix}$, $\lambda_{adv}$ $\lambda_{reg}$  are hyper-parameters for optimization, that control the relative influence of each loss term.

\section{Implementation Details}

\subsubsection*{\textbf{Audio feature extraction}}
Given an audio input, DeepSpeech \cite{hannun2014deep} produces log probabilities of each character (26 alphabets + 3 special characters) corresponding to each audio frame. We use the output of the last layer of the pre-trained DeepSpeech network before applying softmax. We use overlapping audio windows of 16 audio frames (0.04s of audio), where each audio window corresponds to a single video frame. 

\subsubsection*{\textbf{Extraction of facial landmarks}}
We use OpenFace \cite{baltrusaitis2018openface} and face segmentation \cite{yu2018bisenet} to prepare ground truth facial landmarks for training audio-to-landmark prediction network. For a given face image, OpenFace predicts $68$ facial landmarks and uses frame-wise tracking to obtain temporally stable landmarks. But for the lip region, it often gives erroneous prediction, especially for the frames with faster lip movements. To capture an exact lip movement corresponding to the input audio, we need a more accurate method for the ground truth landmark extraction. Hence, we use face segmentation \cite{yu2018bisenet}, which segments the entire face in different regions like hair, eyes, nose, upper lip, lower lip, and rest of the face. We select the upper and lower lip landmark point from the intersection of projected OpenFace landmark points with segmentation boundaries of the lip regions, for a more accurate estimation of lip landmarks.

To prepare ground-truth landmark displacements for training audio-to-landmark prediction network, we impose lip movements on the mean neutral face.
For this, we first align the person-specific landmark $l^p$ with the mean face landmark $l^m_N$ using rigid Procrustes alignment \cite{srivastava2005statistical}. 
Per frame lip displacements from the person-specific neutral face $l^p_N$, are added to the mean neutral face, $l^m_N$ to transfer the motion from person specific landmarks to mean face landmarks, $l^m$. Displacements are scaled with the ratio of person-specific face height-width to mean face height-width before adding to $l^m_N$.
\subsubsection*{\textbf{Landmark Generation from Audio}}
We adopt an encoder-decoder architecture (as shown in Fig. \ref{fig:aud-to-land}) for predicting the landmark displacements. The encoder network consists of four convolution layers with two linear layers in the decoder. We use Leaky ReLU activation after each layer of the encoder network. Input audio feature $a_i$ is reshaped as $\mathbb{R}^{16 \times 1 \times 29}$ to consider the temporal relationship within the window of 16 audio frames. We initialize the decoder layer’s weight with PCA components (that represents 99\% of total variance) computed over landmark displacements of the mean face of training samples. The loss parameters $\lambda_{lmark}$ and $\lambda_{temp}$ have been set to $1$ and $0.5$ respectively based on experimental validation.

\subsubsection*{\textbf{Blink generation network}}
We use RNN architecture to predict a sequence of displacements for each of the landmark points of eye region ($n \times T \times 44$, i.e x, y coordinates of 22 landmarks; n is batch size ) over $T$ timestamps from given noise vector 
$z \sim \mathcal{N}(\mu,\,\sigma^{2})$ of size 10 ($n \times T \times 10$). Fig. \ref{fig:blink-generation} shows network architecture for the blink generation module. Similar to the audio-to-landmark prediction network blink generation network is also trained on landmark displacements. The last linear layer weight is initialized with PCA components (with 99\% variance) computed using eye landmark displacements.  

\subsubsection*{\textbf{Texture Generation from Landmarks}}

The proposed architecture of the texture generator is shown in Fig. \ref{fig:architecture}. 
The current landmark images $L_t$ and the identity landmark image $L_{id}$ images are each encoded using a landmark encoder. The difference in encoded landmark features is concatenated with the input identity image $I_{id}$ and fed to an encoder-decoder architecture, which generates attention map $att_t$ and color map $C_t$. The generated image $I_{t}$ is passed to a discriminator network, which determines if the generated image is real or fake. The encoder-decoder architecture of the generator network is built upon a variation of \cite{pumarola2018ganimation} which uses facial action units to generate attention for facial expression generation. The discriminator network is based on the PatchGan architecture \cite{isola2017image} with batch normalization replaced by instance normalization similar to \cite{pumarola2018ganimation} for greater training stability.
The improved stability of LSGAN training \cite{mao2017least} along with regularization of attention map helped us in achieving stable adversarial training as the problem of vanishing gradients in the regular GAN training can adversely effect learning of attention and color maps. 
We use Adam optimizer with learning rate of $0.0001$, $beta1=0.5$, $beta2=0.999$ and training batch size of $16$. During training the loss hyper-parameters have been set to $\lambda_{pix}=100$, $\lambda_{adv}=0.5$ and $\lambda_{reg}=0.2$ by experimental validation on a validation set. The adversarial loss and regularization loss parameters have been chosen to prevent saturation of the attention map while maintaining the sharpness of the texture of the generated images.

Our network is trained on a NVIDIA Quadro GV100 GPU. Training of audio-to-landmark, blink, and landmark-to-image generation networks take around 6 hours, 3 hours, and 2 days respectively. We use PyTorch for the implementation of the above mentioned networks.

\begin{figure*}[ht]
\begin{subfigure}[t]{\textwidth}
\center
\includegraphics[width=0.7\textwidth]{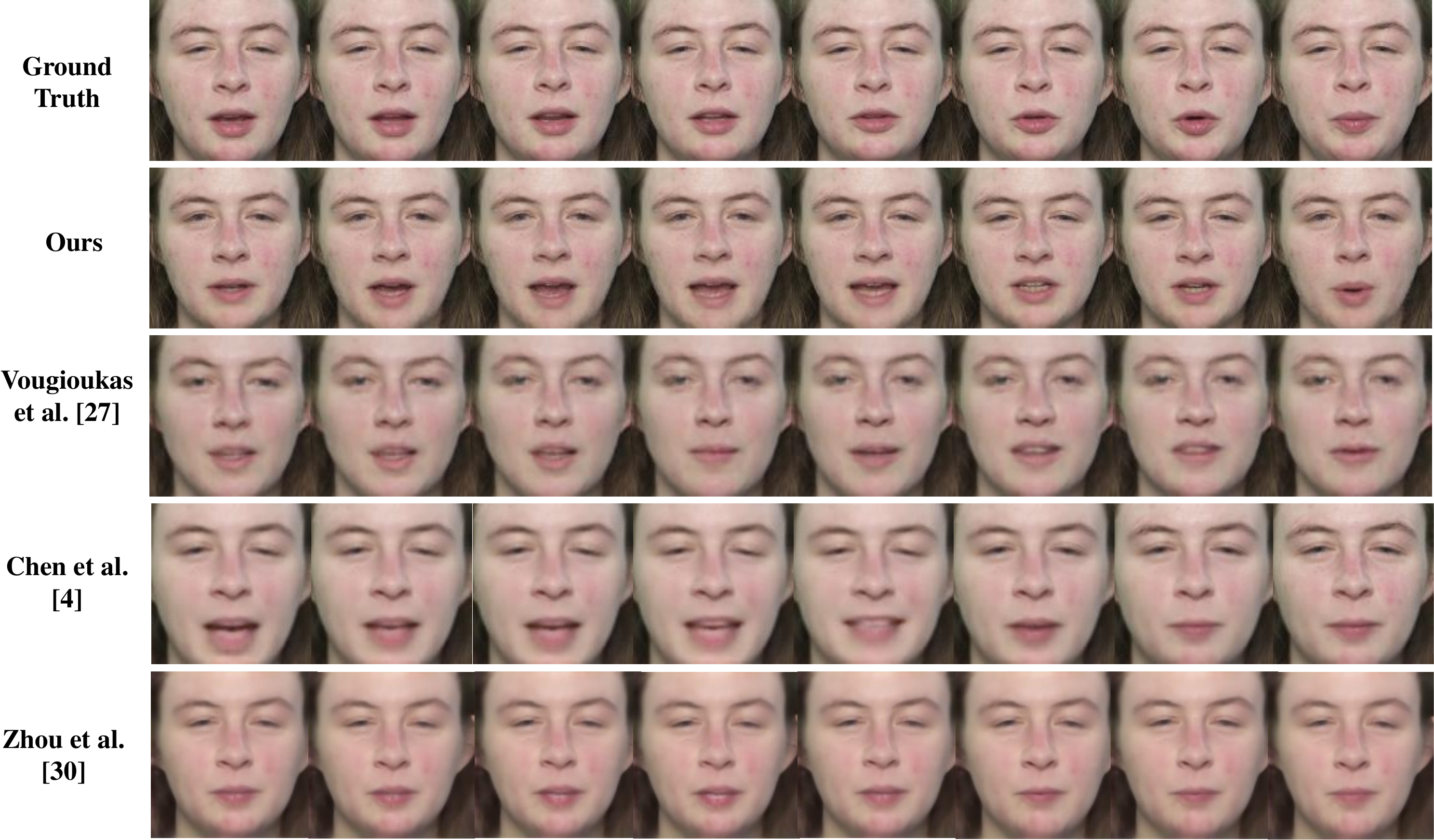}
\end{subfigure}
\par \smallskip
\begin{subfigure}[t]{\textwidth}
\center
\includegraphics[width=0.7\textwidth]{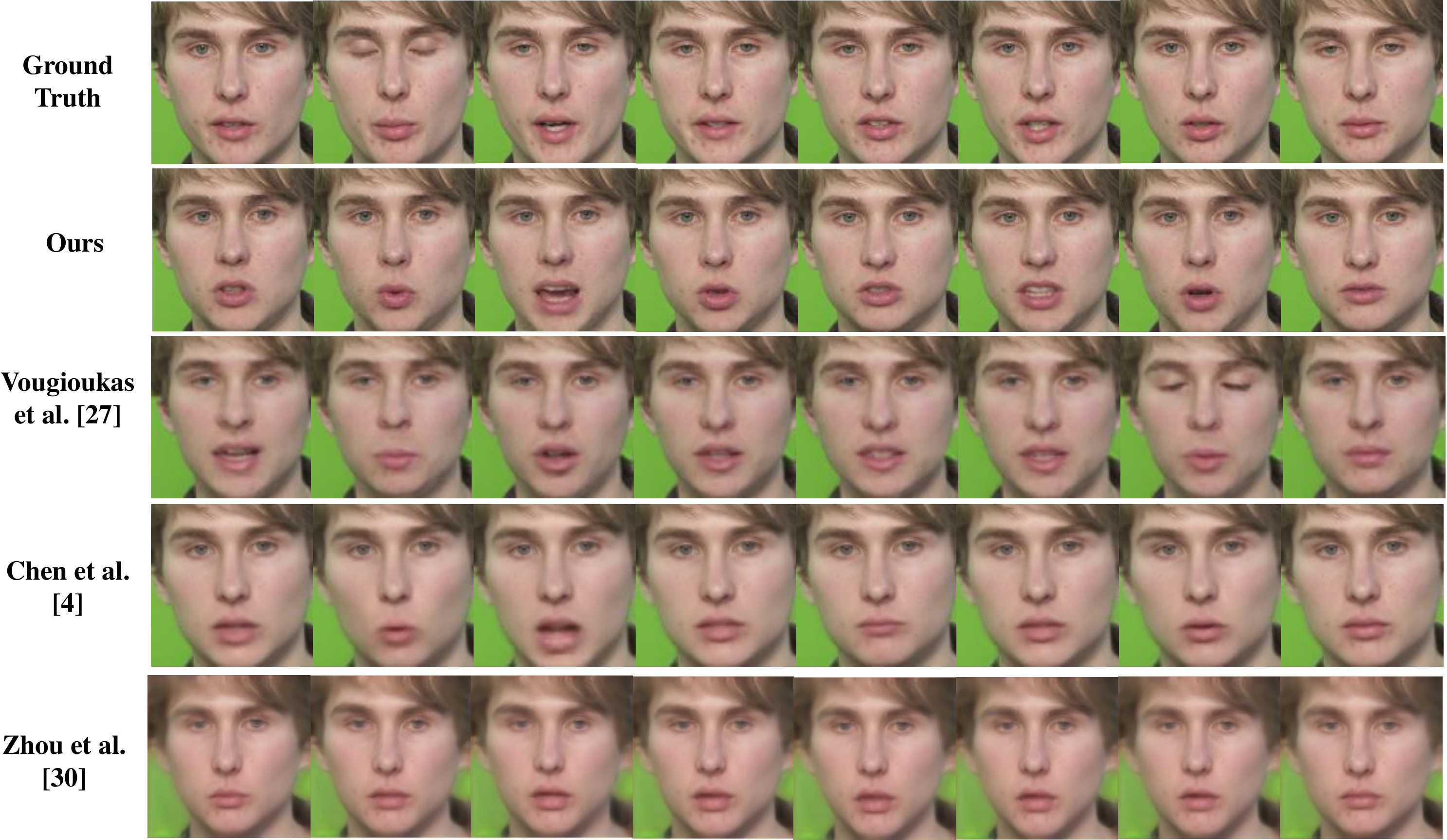}
\end{subfigure}
\caption{Results on TCD-TIMIT dataset: Our generated texture is sharper, especially the texture of the mouth and teeth are visibly more distinct compared to Chen et al. \cite{chen2019hierarchical}, Vougioukas et al. \cite{vougioukas2019realistic} and Zhou et al.  \cite{zhou2019talking} and also our generated motion is better than Zhou et al. \cite{zhou2019talking}. In contrast to Vougioukas et al. \cite{vougioukas2019realistic} and Zhou et al. Zhou et al. \cite{zhou2019talking} our synthesized face retains the texture from the input identity image in the regions of the face not undergoing motion, resulting in our improved identity-preservation. }
\label{fig:TIMIT_qualitative}
\end{figure*}

\begin{figure*}[tb]
\begin{subfigure}[t]{\textwidth}
\center
\includegraphics[width=0.7\textwidth]{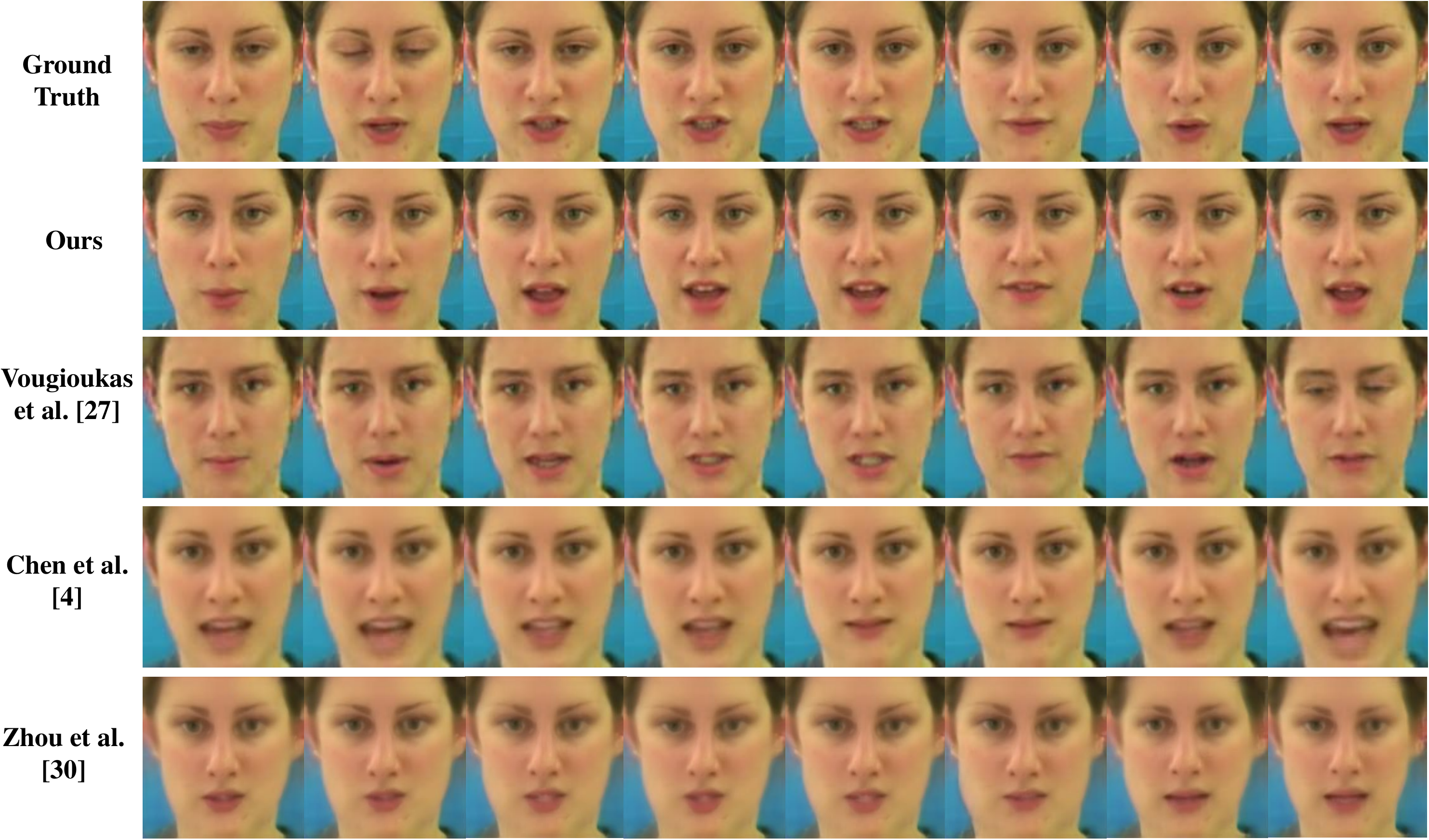}
\end{subfigure}
\par \smallskip
\begin{subfigure}[t]{\textwidth}
\center
\includegraphics[width=0.7\textwidth]{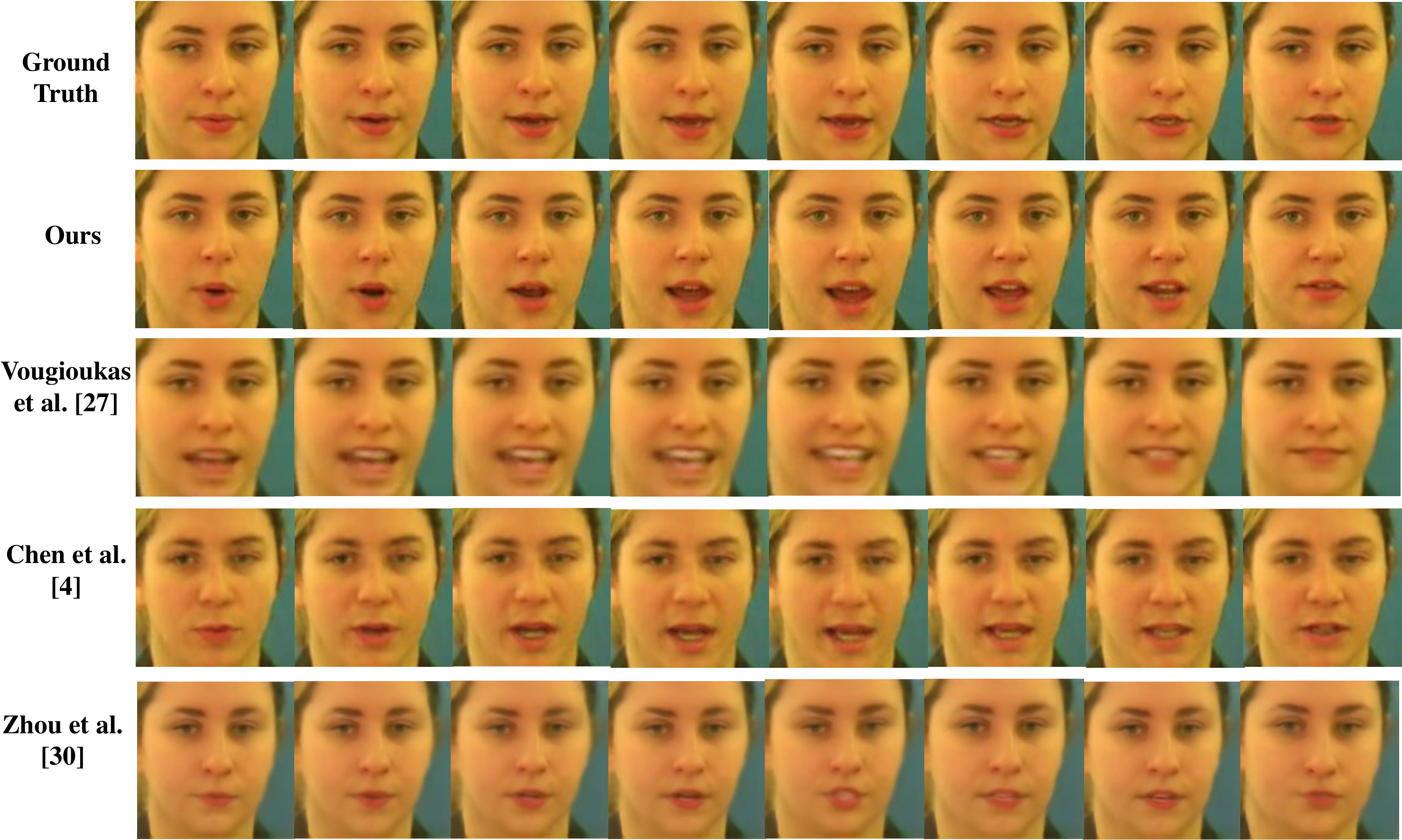}
\end{subfigure}
\caption{Results on GRID dataset: Our generated images contain sharper and more distinctive mouth texture, plausible mouth shapes, and better preservation of identity compared to Chen et al. \cite{chen2019hierarchical}, Vougioukas et al. \cite{vougioukas2019realistic} and Zhou et al. \cite{zhou2019talking}. Vougioukas et al. \cite{vougioukas2019realistic} fails to accurately preserve the identity information of the target in the synthesized images, Chen et al. \cite{chen2019hierarchical} and Zhou et al. \cite{zhou2019talking} contain some implausible mouth shapes. }
\label{fig:GRID_qualitative}
\end{figure*}
\begin{figure*}[t]
\centering
\includegraphics[width=0.9\textwidth]{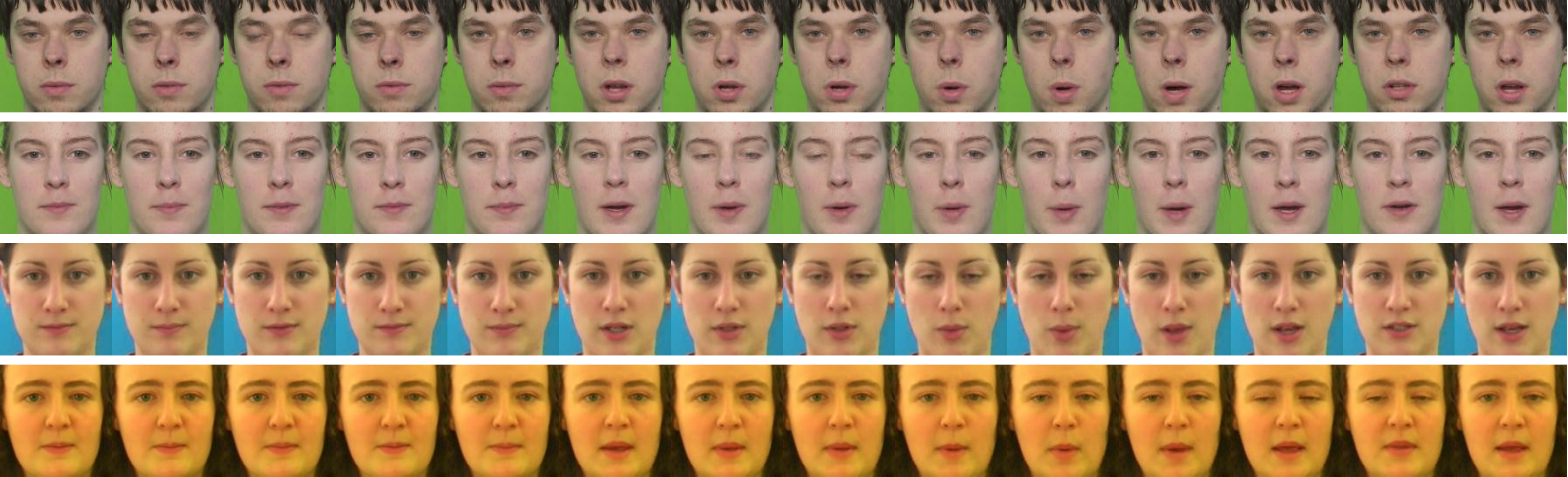}
\caption{Our generated animation of different identities synchronized with the the same speech input, containing spontaneous generation of eye blinks. }
\label{fig:results_blinks_same_audio}
\end{figure*}

\section{Experiments}

The proposed model is trained and evaluated on the benchmark datasets GRID \cite{cooke2006audio} and TCD-TIMIT\cite{harte2015tcd}. 
The GRID dataset consists of $33$ speakers, each uttering $1000$ short sentences, but the words belong to a limited dictionary. The TCD-TIMIT dataset consists of $59$ speakers uttering approximately $100$ sentences each from the TIMIT corpus, with long sentences that contain much more phonetic variability than the GRID dataset. We use the same training-testing data split for the TCD-TIMIT and GRID datasets as in \cite{vougioukas2019realistic}. 

\subsection{Metrics}

The following metrics have been used for quantitative evaluation of our results:

- Image reconstruction quality metrics, PSNR (peak signal-to-noise ratio), and SSIM (structural similarity). 

- Image sharpness metric CPBD (Cumulative probability blur detection)\cite{narvekar2009no} to detect the amount of blur in synthesized image. 


- Landmark synchronization metric LMD (landmark distance) \cite{chen2018lip} to measure the accuracy of audio-visual synchronization. 

Higher values of CPBD, PSNR, and SSIM indicated better quality of image generation while lower values of LMD indicate better audio-visual synchronization.

\subsection{Results}

Our results have been compared both qualitatively and quantitatively with recent state-of-the-art methods. A user study has also been carried out for subjective evaluation of our method. 

\subsubsection{Qualitative Results}

Qualitative comparison of our results have been carried out with the recent state-of-the-art methods of Chen et al. \cite{chen2019hierarchical}, Vougioukas et al. \cite{vougioukas2019realistic} and Zhou et al. \cite{zhou2019talking}.
The comparative results on TCD-TIMIT and GRID dataset are shown in Fig. \ref{fig:TIMIT_qualitative} and \ref{fig:GRID_qualitative} respectively. The results indicate that our proposed method is able to generate facial animation sequences that are superior in terms of image quality, identity preservation and generation of plausible mouth shapes. Our generated images contain sharper texture and are better at preserving the identity-related facial texture of the target subjects compared to Vougioukas et al. \cite{vougioukas2019realistic} and Zhou et al. \cite{zhou2019talking} due to our attention-based texture generation with the help of landmarks, which helps to retain the identity information from the input identity image. 
Compared to Chen et al. \cite{chen2019hierarchical}, our generated face images have less blur and more distinctive texture in the mouth region and plausible mouth shapes. This is because of our two-step learning of person-specific facial landmarks, and texture generation using LSGAN and attention map regularization. Unlike Chen et al. \cite{chen2019hierarchical} and Zhou et al. \cite{zhou2019talking}, our face animation method can generate spontaneous eye blinks, as shown in Fig. \ref{fig:results_blinks_same_audio}.

\subsubsection{Quantitative Results}
In this section, we present a quantitative evaluation of our method compared with the recent methods  \cite{chen2019hierarchical,vougioukas2019realistic}. 
Table \ref{tab:metrics} shows the metrics computed using our trained models on GRID and TCD-TIMIT datasets respectively. Our results indicate better image reconstruction quality (higher PSNR and SSIM), sharper texture (higher CPBD) and improved audio-visual synchronization (lower LMD) than the state-of-the-art methods\cite{chen2019hierarchical,vougioukas2019realistic}.
\begin{table}[tb]
\noindent\resizebox{\columnwidth}{!}{
 \begin{tabular}{||c | c | c | c | c | c | c ||} 
 \hline
Dataset &Method & PSNR & SSIM & CPBD & LMD\\ [0.5ex] 
 \hline\hline
 \rule{0pt}{3ex} 
\multirow{2}{*}{TCD-TIMIT}& Ours & \textbf{26.153} & \textbf{0.818} & \textbf{0.386} & \textbf{2.39} \\
 &Vougioukas et al. \cite{vougioukas2019realistic} & 24.243 & 0.730 & 0.308  & 2.59 \\
  &Chen et al. \cite{chen2019hierarchical} & 20.311 & 0.589 & 0.156 &  2.92 
  \\[1ex] 
  
 \hline
  \rule{0pt}{3ex} 
 \multirow{2}{*}{GRID}& Ours & \textbf{29.305} & \textbf{0.878} &  \textbf{0.293}  & \textbf{1.21}\\
 &Vougioukas et al. \cite{vougioukas2019realistic} & 27.100 & 0.818 & 0.268 & 1.66 \\
  &Chen et al. \cite{chen2019hierarchical} & 23.984 & 0.7601 & 0.0615 & 1.59 
  \\[1ex] 
  \hline 
\end{tabular} 
}

\caption{Quantitative evaluation results. We evaluate the methods of Chen et al. \cite{chen2019hierarchical}
Vougioukas et al.\cite{vougioukas2019realistic} 
on our test data using their respective pre-trained models which are publicly available. Our train-test split is same as that of Vougioukas et al.\cite{vougioukas2019realistic}. }
\label{tab:metrics}
\end{table}

\begin{figure}[t]
    \centering
\includegraphics[width=0.6\columnwidth]{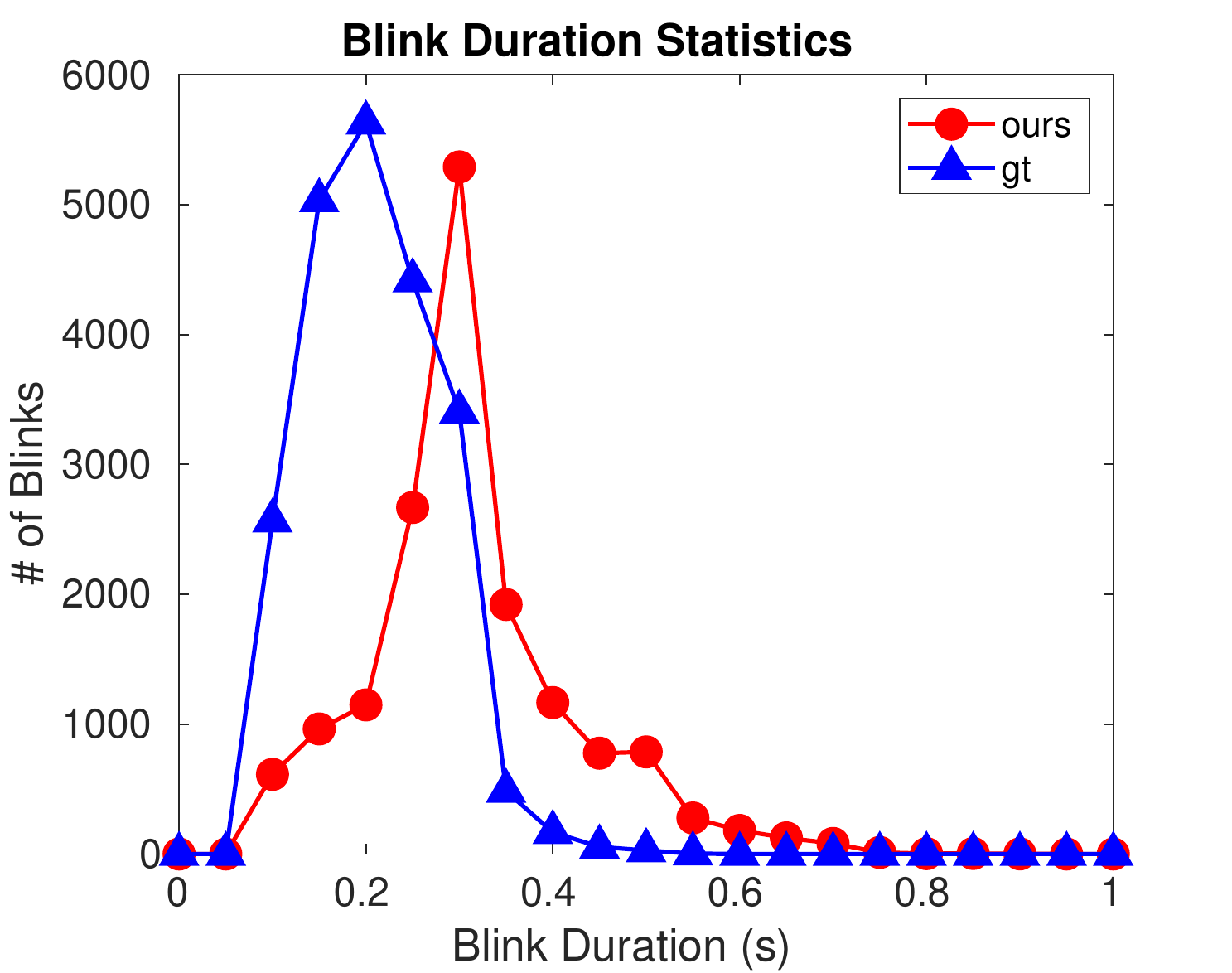}
    \caption{Blink duration in synthesized videos compared to ground-truth.}
    \label{fig:blink_stats}
\end{figure}

We also evaluate the performance of our blink generation network by comparing the characteristics of predicted blinks with blinks present in ground-truth videos.
Fig. \ref{fig:blink_stats} shows the comparison of the distributions of blink duration for around 11,000 synthesized (red) and ground-truth (blue) videos (from GRID and TCD-TIMIT datasets). The average blink duration per video obtained from our method is similar to that of ground-truth. Our method produces $0.3756$ blinks/s and $0.2985$ blinks/s for GRID and TCD-TIMIT datasets respectively which is similar to the average human blink rate, that varies between $0.28-0.4$ blinks/s \cite{vougioukas2019realistic}. Also, our method shows an average of $0.5745s$ inter-blink duration, which is similar to ground-truth videos with duration $0.4601s$. Hence, our method is able to produce realistic blinks.

\begin{table}[tb]
\begin{center}
    
\resizebox{0.8\columnwidth}{!}{
 \begin{tabular}{||c | c | c | c ||} 
 \hline
Method & PSNR & SSIM & CPBD\\ [0.5ex] 
 \hline\hline
  $L_{pix}$ & 25.874 & 0.813 & 0.366\\
  $L_{pix}+L_{adv}$ &25.951 & 0.814 & 0.373 \\
  $L_{pix}+L_{adv}+L_{reg}$ &\textbf{26.153} & \textbf{0.818} & \textbf{0.386}\\
  \hline 
\end{tabular} 
}
\caption{Ablation study of the objective function in Eq. \ref{eqn:textureobjectivefunction} on the TCD-TIMIT dataset.}
\label{tab:ablation}
\end{center}
\end{table}

\begin{table}[tb]
\begin{center}
\resizebox{0.9\columnwidth}{!}{%
\begin{tabular}{||c | c | c | c ||} 
 \hline
Method & TCD-TIMIT & GRID & Average \\ [0.5ex] 
 \hline\hline
 Ours & \textbf{6.40} & \textbf{7.69} & \textbf{7.05}\\
  Vougioukas et al. \cite{vougioukas2019realistic} & 6.29 & 6.51 & 6.4\\
  Chen et al. \cite{chen2019hierarchical} &4.67 & 4.5 & 4.59 \\
  \hline 
\end{tabular}
}
\caption{User study results. Scores range from 0-10 (Higher scores indicate more realistic face animation) }
 \label{tab:userstudy}
\end{center}
\end{table}

\subsubsection{Ablation Study}
\begin{figure}[t]
\center
\includegraphics[ width=\columnwidth]{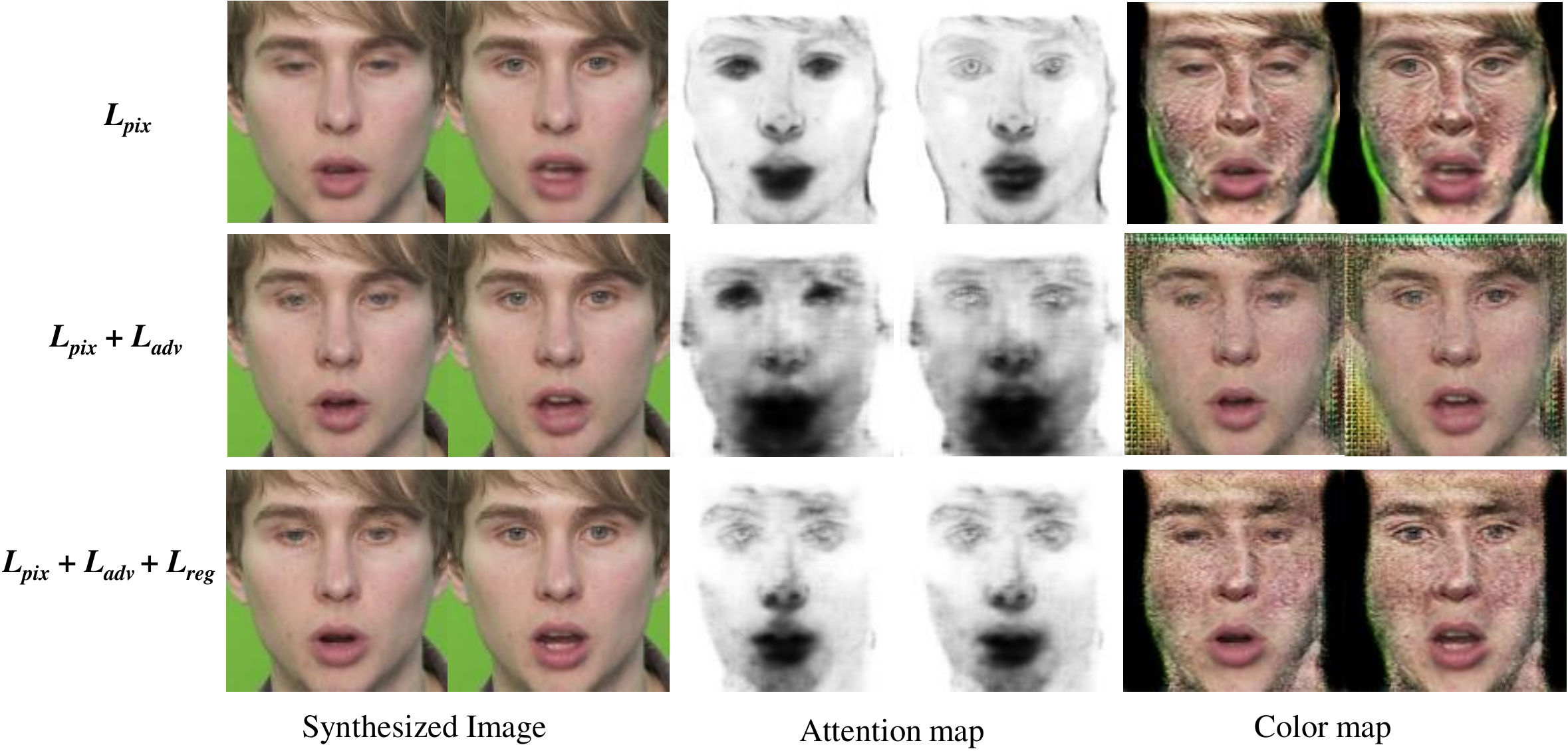}

\caption{Training the network using only generator loss $L_{pix}$ without the discriminator, results in blurry texture generation in the mouth region of the color map. Adding the discriminator and the adversarial loss (row marked $L_{pix} + L_{adv}$) makes the generated mouth texture sharper in the color map, however the attention map indicates motion for the entire face resulting in blur in the final synthesized image, especially noticeable in the mouth region. Adding the regularization loss (row marked $L_{pix} + L_{adv} + L_{reg}$) results in the attention map having low values mostly in regions of motion, hence the synthesized image contains sharper and more distinct mouth texture. }  
\label{fig:ablation}
\end{figure}

We present an ablation study on a validation set from TCD-TIMIT, for different losses (Eq. \ref{eqn:textureobjectivefunction}) used for training our landmark-to-image generation network. This helps to understand the significance of using adversarial training and regularization. The metrics are summarized in Table \ref{tab:ablation} and generated images are shown in Fig. \ref{fig:ablation}. The results indicate that our texture generation network trained using a combination of $L_1$ pixel loss, adversarial loss, and regularization yields the best outcome. 

\subsubsection{User Study}
A user study has also been carried out to
evaluate the realism of our facial animation results. 26 participants have rated 30 videos with a score between 0-10 (higher score indicates more realistic). Out of the 30 videos, 10 videos are selected from each of the following methods - Ours, Vougioukas et al. \cite{vougioukas2019realistic} and Chen et al. \cite{chen2019hierarchical}. For each method, 5 videos are selected from each of the datasets, GRID, and TCD-TIMIT. Table \ref{tab:userstudy} summarizes the outcome of the user study, which indicates higher realism for the synthesized videos generated by our method. 
As per the feedback from the participants, our sharper images, better identity preservation over the videos, and the presence of realistic eye blinks helped us achieve higher scores indicating improved realism compared to state-of-the-art methods.

\section{Conclusion}
In this paper, we propose an efficient pipeline for generating realistic facial animation from speech. Our method produces accurate audio-visual synchronization, plausible mouth movement along with identity preservation and also renders natural expression like eye blinks. Our results indicate a significant improvement over the state-of-the-art methods in terms of image quality, speech-synchronization, identity-preservation, and overall realism, as established by our qualitative, quantitative and user study results. We attribute this to our segregated learning of motion and texture, two-stage learning of person-independent and person-specific motion, generation of eye blinks, and the use of attention to retain identity information. In future, we would like to generate a greater variety of spontaneous human expressions and head movements to make the animation appear more realistic.
\section{Acknowledgement}
We would like to acknowledge Prof. Angshul Majumdar from IIIT Delhi, India, for helping us gain the access of TCD-TIMIT dataset for our research.
\bibliographystyle{ieee}

\end{document}